\documentclass{article}

\PassOptionsToPackage{numbers}{natbib}

\usepackage[preprint]{neurips_2024}

\usepackage[utf8]{inputenc} 
\usepackage[T1]{fontenc}    
\usepackage[colorlinks=true, citecolor=violet]{hyperref}
\usepackage{url}            
\usepackage{booktabs}       
\usepackage{amsfonts}       
\usepackage{nicefrac}       
\usepackage{microtype}      
\usepackage{xcolor}         
\usepackage{graphicx}
\usepackage{multirow}
\usepackage{ltablex}
\keepXColumns
\usepackage{tabularx}
\usepackage{caption}
\usepackage{etoolbox}

\makeatletter

\title{Improving Social Determinants of Health Documentation in French EHRs Using Large Language Models}

\author{%
  Adrien Bazoge$^{1,2}$ \quad Pacôme Constant dit Beaufils$^{1,3,4}$ \quad Mohammed Hmitouch$^{1,2}$ \\ 
  \textbf{Romain Bourcier}$^{3}$ \quad \textbf{Emmanuel Morin}$^{2}$ \quad \textbf{Richard Dufour}$^{2}$ \quad \textbf{Béatrice Daille}$^{2}$ \\
  \textbf{Pierre-Antoine Gourraud}$^{1}$ \quad \textbf{Matilde Karakachoff}$^{1}$ \\
  $^1$Data Clinic, University Hospital of Nantes, France\\
  $^2$Nantes Université, École Centrale Nantes, CNRS, LS2N, France\\
  $^3$Department of Neuroradiology, University Hospital of Nantes, Thorax Institute, France\\
  $^4$Nantes Université, CNRS, INSERM, Thorax Institute, France\\
}

\begin{document}

\maketitle

\begin{abstract}
  Social determinants of health (SDoH) significantly influence health outcomes, shaping disease progression, treatment adherence, and health disparities. However, their documentation in structured electronic health records (EHRs) is often incomplete or missing. This study presents an approach based on large language models (LLMs) for extracting 13 SDoH categories from French clinical notes. We trained Flan-T5-Large on annotated social history sections from clinical notes at Nantes University Hospital, France. We evaluated the model at two levels: (i) identification of SDoH categories and associated values, and (ii) extraction of detailed SDoH with associated temporal and quantitative information. The model performance was assessed across four datasets, including two that we publicly release as open resources.
  The model achieved strong performance for identifying well-documented categories such as living condition, marital status, descendants, job, tobacco, and alcohol use (F1 score > 0.80). Performance was lower for categories with limited training data or highly variable expressions, such as employment status, housing, physical activity, income, and education. Our model identified 95.8\% of patients with at least one SDoH, compared to 2.8\% for ICD-10 codes from structured EHR data. Our error analysis showed that performance limitations were linked to annotation inconsistencies, reliance on English-centric tokenizer, and reduced generalizability due to the model being trained on social history sections only. These results demonstrate the effectiveness of NLP in improving the completeness of real-world SDoH data in a non-English EHR system.

\end{abstract}

\section{Introduction}

Despite a continuous increase in overall life expectancy, social inequalities in health persist and are widening throughout the life course~\cite{merville2024unpacking,chetty2016association}. Medical progress can prolong the life expectancy of individuals with severe diseases, but care management cannot be disconnected from the socio-economic environments where patients live~\cite{marmot2005social}. The outcomes of chronic diseases are shaped by a combination of behaviors and exposures, resulting in a complex socio-biological process that determines both individual and societal health status~\cite{halfon2014lifecourse}.
Social determinants of health (SDoH) are the conditions in which people grow, live, work, and age that influence their quality of life and health~\cite{halfon2014lifecourse}. These determinants encompass a broad range of socioeconomic factors, including family support, employment status, and education, as well as health-related behaviors, such as substance use and physical activity~\cite{alderwick2019meanings}. 
In addition to shaping behaviors, socioeconomic conditions also determine individuals’ exposure to environmental risks, such as air pollution, noise, and extreme weather events, which further impact health outcomes~\cite{nettle2010there}. Together, social and behavioral factors are major drivers of health disparities, contributing to 47\% and 34\% of patient outcomes, respectively~\cite{hood2016county}. In clinical settings, SDoH are documented in the electronic health records (EHR) through both structured data (e.g., coded fields) and unstructured data (e.g., clinical notes)~\cite{wang2021documentation}. However, unstructured clinical notes provide more intricate and detailed representations of SDoH than structured data~\cite{chen2011multi}. To enable large-scale secondary use of EHR data, there is an increasing need for automated methods to extract and structure patient SDoH, enhancing our ability to study the impact of social inequalities on health~\cite{bazoge2023applying}. In this context, studying SDoH in clinical sciences is essential for understanding how factors like income, education, housing, and social environments shape health outcomes beyond technico-biological variables. Integrating insights from SDoH into clinical practice highlights opportunities for early intervention, holistic patient care, and addressing health inequities at their root. Furthermore, recognizing the actionable nature of SDoH enables the development of evidence-based public health policies that target structural barriers and improve population health at scale~\cite{hacker2022social}.

In recent years, automatic extraction of SDoH has been widely studied in the English language using natural language processing (NLP)~\cite{patra2021extracting}. Progress has been accelerated by the dissemination of annotated corpora and shared task challenges, for example the i2b2 NLP Smoking Challenge on identifying patients’ smoking status with a corpus of 502 clinical notes~\cite{uzuner2008identifying} and the 2022 n2c2 exploring the extraction of SDoH from 4,405 social history sections from clinical notes, including substance use (alcohol, drugs, and tobacco), employment and living conditions~\cite{lybarger20232022}. 
Most subsequent studies have focused on U.S. hospital settings, showing that NLP methods applied to unstructured data from EHRs can reliably identify key social risk factors~\cite{wray2021examining,feller2018towards,afshar2019natural,stemerman2021identification,patra2025extracting}. 
In terms of coverage, the most commonly addressed SDoH are smoking status, substance abuse (alcohol and drug) and housing instability~\cite{patra2021extracting}, which are well known for their impact on health, and thus are well documented in EHRs~\cite{hood2016county}. In contrast, other SDoH, such as education, employment status, social support and isolation, remain underexplored and present ongoing challenges for NLP systems~\cite{patra2021extracting}.
To address SDoH identification and extraction, a range of NLP methods have been developed. Early approaches relied on rule-based systems and keyword matching, offering high precision but limited generalizability~\cite{conway2019moonstone}. Subsequently, semantic word embedding methods such as word2vec~\cite{mikolov2013distributed} enabled more nuanced lexical representations, supporting downstream machine learning classifiers for SDoH identification~\cite{bejan2018mining,topaz2019extracting,gundlapalli2013using,rouillard2022evaluation,feller2020detecting}. More advanced approaches have leveraged deep learning architectures, CNNs~\cite{han2022classifying}, LSTMs~\cite{han2022classifying}, and transformer-based models such as BERT~\cite{han2022classifying,yu2022study,richie2023extracting,gong2025boosting}, which have demonstrated improved performance in extracting contextually rich SDoH information.

Recent studies have also shown the potential of large language models (LLMs) for the identification and classification of SDoH in clinical notes. Decoder-only models such as GPT-4~\cite{achiam2023gpt} have been used in zero- and few-shot settings~\cite{guevara2024large}, while encoder-decoder models such as Flan-T5~\cite{chung2024scaling} have been applied for generative SDoH extraction tasks, and are the current state-of-the-art approaches for SDoH extraction~\cite{lybarger20232022,patra2025extracting,gong2025boosting,guevara2024large,romanowski2023extracting,keloth2025social}. However, most of available work are focused on English language, while resources for other languages remain scarce~\cite{patra2021extracting,chen2020social}. The extraction of smoking status from clinical narrative texts has been studied in Spanish~\cite{figueroa2014identifying}, Finnish~\cite{karlsson2021impact}, Swedish~\cite{caccamisi2020natural}, and in a Korean-English bilingual setting~\cite{bae2021keyword}, but none of the underlying corpora are available. To our knowledge, the extraction of SDoH in French clinical texts has not been addressed. 

In this work, we propose a sequence-to-sequence approach based on a large language model for extracting SDoH from French clinical texts. Our study focuses on 13 SDoH categories: living condition, marital status, descendants, employment status, occupation, tobacco use, alcohol use, drug use, housing, education, physical activity, income, and ethnicity/country of birth. To support model development and evaluation, we constructed and manually annotated four datasets consisting of social history sections from clinical notes. Two of these datasets are publicly released to promote reproducibility and support the development of new methods for SDoH extraction in French.

\newpage

\section{Methods}

\subsection{Data}

\begin{figure}[!h]
  \centering
  \includegraphics[width=0.50\textwidth]{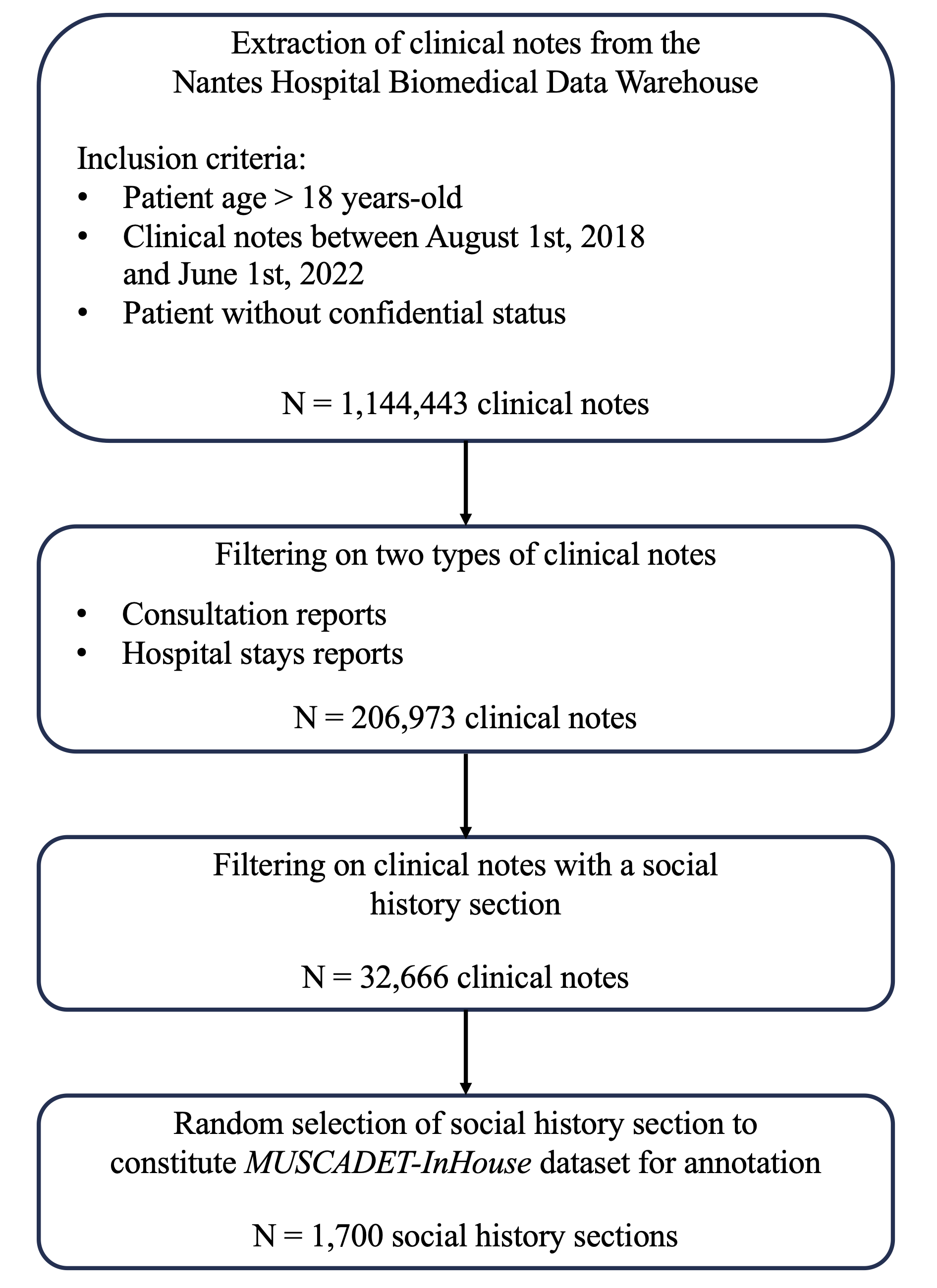}
  \caption{MUSCADET-InHouse corpus construction flow-chart.}
  \label{fig:figure1}
\end{figure}

To train and evaluate the proposed SDoH model, we used four datasets: MUSCADET-InHouse, MUSCADET-Synthetic, UW-FrenchSDOH and InHouse Tuberculosis/ALS.
\paragraph{MUSCADET-InHouse} This dataset was obtained from clinical notes from the Nantes biomedical data warehouse (NBDW), as summarized in Figure~\ref{fig:figure1}. The NBDW encompasses nearly 1.5 million patients who received care at the Nantes University Hospital, over the past 20 years.  It includes different dimensions of patient-related data: structured data (e.g. \textit{Classification Commune des Actes Médicaux} billing codes, a French coding system of clinical procedures; ICD-10 codes; laboratory results; drug administrations), and unstructured data such as outpatient and inpatient clinical notes, radiology and operative reports~\cite{karakachoff2024implementing}. The NBDW was authorized by the French authority of data protection (\textit{Commission Nationale de l’Informatique et des Libertés}) (Registration code n°920242). The present study is compliant with French regulatory and General Data Protection Regulation requirements, including informed consent. A total of 1,144,443 clinical notes were selected from the NBDW according to the following inclusion criteria: age $\geq$ 18 years, the presence of clinical notes within the NBDW between August 1, 2018, and June 1, 2022. The non-inclusion criterion was patient opposition to data reuse. We then focused on semi-structured clinical notes containing predefined sections (e.g., 'History', 'Medications', 'Social History', etc.), with a particular emphasis on two categories: consultation reports and hospital stay reports. These two types of notes span multiple medical specialties, resulting in a total of 206,973 clinical notes covering diverse patient profiles. The clinical notes selected in the previous step were filtered to retain only those containing a social history section, for a total of 32,666 clinical notes. The social history section was extracted using a rule-based approach. Finally, 1,700 social history sections were randomly selected to constitute our corpus for annotation. This dataset was randomly divided into training (70\%), validation (10\%) and test (20\%) sets for our experiments.

To assess the generalization capabilities of our model, we constructed three external test datasets. For reproducibility, we introduce two open-source datasets: MUSCADET-Synthetic and UW-FrenchSDOH.
\paragraph{MUSCADET-Synthetic} This dataset comprises synthetic social history section texts written by a physician. These synthetic documents follow the template of real medical records but were entirely written from scratch, ensuring that they do not reference any real patient. The corpus includes 340 documents, matching the test set size of MUSCADET-InHouse.

\paragraph{UW-FrenchSDOH} This dataset is an automatically translated version of an existing dataset from the University of Washington (UW)~\cite{yetisgen2016automatic,yetisgen2017automatic}. It consists of 364 social history sections collected from MTSamples. The dataset was translated into French using GPT-4o (\texttt{gpt-4o-2024-11-20}) and manually corrected during annotation. 

\paragraph{InHouse Tuberculosis and ALS} Since other datasets focus only on social history sections, it was essential to assess the model’s effectiveness in a broader clinical context, particularly on non-SDoH texts, to determine its propensity for false positives. To this end, we applied the model to full clinical notes in two use cases, focusing on patients hospitalized for tuberculosis or amyotrophic lateral sclerosis (ALS). These diseases were selected due to the significant impact of SDoH on their outcomes~\cite{koltringer2023social,hargreaves2011social,alonso2010association}. Both groups of patients and related clinical notes were selected on ICD-10 criteria: A15-A19 for tuberculosis and G12.2 for ALS. A total of 1,186 patients were identified for tuberculosis and 647 for ALS. We included the first clinical note recorded for each patient visit associated with the respective ICD-10 code as the principal diagnosis. For each disease, 200 clinical notes were fully annotated to serve as a test set, for a total of 400 clinical notes.

\subsection{Annotation Scheme}

The annotation scheme was designed to provide a broad coverage of SDoH, with a fine-grained description of the determinants. The annotation scheme includes entities, attributes, and relations between entities. 

The annotation scheme comprises 25 entities related to SDoH, covering 13 SDoH categories (living condition, marital status, descendants, employment status, job, tobacco use, alcohol use, drug use, housing, education, physical activity, income and ethnicity/country of birth) and 6 entities related to relations (StatusTime, History, Duration, Amount, Frequency, Type). SDoH entities are either span-only (Job, Income, Education, Ethnicity, Alcohol, Tobacco, Drug) or labeled (Living, MaritalStatus, Descendants, Employment, Housing, PhysicalActivity), while all relation entities are span-only. Table~\ref{tab:entities} presents all entities. 

\setcounter{table}{0}
{\scriptsize
{\normalsize \captionof{table}{Entities description.}}
\protected@edef\@currentlabel{\arabic{table}}
\label{tab:entities}
\begin{tabularx}{\textwidth}{lXXX}
\toprule
\textbf{Entity} & \textbf{Definition} & \textbf{Examples} & \textbf{Examples translated into \mbox{English}} \\
\midrule
\endfirsthead

\toprule
\textbf{Entity} & \textbf{Definition} & \textbf{Examples} & \textbf{Examples translated into \mbox{English}} \\
\midrule
\endhead

\midrule
\multicolumn{4}{r}{\textit{Continued on next page}} \\
\bottomrule
\endfoot

\bottomrule
\endlastfoot

    Living\_Alone & The patient is described as living alone.  & \textit{Vit seul}  & \textit{Living alone}   \\
    \cmidrule(r){1-4}
    Living\_WithOthers & The patient is described as living with other people. This doesn’t include animals but includes nursing home.  & \textit{Vit avec sa soeur; Vit avec ses parents; Vit avec son épouse et ses enfants ; Vit en EHPAD}  & \textit{Lives with his/her sister; Lives with his/her parents; Lives with his wife and children; Lives in nursing home}   \\
    \cmidrule(r){1-4}
    MaritalStatus\_Single & The patient described as being single (i.e. no regular partner).  & \textit{Célibataire sans enfant; Pas de conjoint}  & \textit{Single without children; No partner}   \\
    \cmidrule(r){1-4}
    MaritalStatus\_InRelationship & The patient is described as married, registered partnership or with a regular partner.  & \textit{Mariée; Vit en concubinage; Vit avec sa compagne; Pacsé}  & \textit{Married ; Lives with partner; Lives with his girlfriend; Civil union}   \\
    \cmidrule(r){1-4}
    MaritalStatus\_Divorced & The patient is described as divorced or separated from is regular partner (except from non-regular partnership).  & \textit{Divorcé; Plus de contact avec le père de ses enfants}  & \textit{Divorced; No more contact with the father of her children}   \\
    \cmidrule(r){1-4}
    MaritalStatus\_Widowed & The patient is described as widowed. & \textit{Veuf; Veuve}  & \textit{Widower, Widow}   \\
    \cmidrule(r){1-4}
    Descendants\_Yes & The patient has descendants. & \textit{Deux filles; 1 fils adopté; Trois petits-enfants; Courses faites par sa fille}  & \textit{Two daughters; 1 adopted son; Three grandchildren; Shopping done by daughter}   \\
    \cmidrule(r){1-4}
    Descendants\_No & The patient has no descendants. & \textit{Pas d’enfant; Sans enfant; Enceinte de son premier enfant}  & \textit{No children; Childless; Pregnant with first child}   \\
    Job & All the patient’s jobs, previous and actual. & \multirow{2}{*}{\begin{minipage}[t]{0.20\columnwidth}
         \raggedright \textit{Courtier dans le textile ; Travaille dans le bâtiment; Soudeur}
         \end{minipage}}   & \multirow{2}{*}{\begin{minipage}[t]{0.20\columnwidth}
         \raggedright \textit{Textile broker; Construction worker; Welder}
         \end{minipage}}   \\
    \cmidrule(r){1-2}     
    Last\_job & The patient’s latest job. &   &    \\
    \cmidrule(r){1-4}
    Employment\_Working & Any documentation that the patient is currently working. It includes short/finite period of time off from work as sick leave, sabbatical, maternity leave or only information about job. & \textit{Travaille toujours; En reconversion professionnelle ; Agent de ménage ; Maître d’œuvre}  & \textit{Still working; Retraining; Housekeeper; Prime contractor}   \\
    \cmidrule(r){1-4}
    Employment\_Unemployed & The patient is described as unemployed. & \textit{Vient de terminer son CDD ; Il est au chômage}  & \textit{Has just finished his fixed-term contract; He is unemployed}   \\
    \cmidrule(r){1-4}
    Employment\_Student & The patient is described as a student. & \textit{Actuellement lycéen; \dots en première année de BTS ; Patient étudiant}  & \textit{Currently in high school; \dots in first year of associated degree ; Student patient}   \\
    \cmidrule(r){1-4}
    Employment\_Pensioner & The patient is described as pensioner. It also includes mentions such as “former” + job references, without explicit mention of pensioner. & \textit{Retraité; A la retraite; Ancien magasinier}  & \textit{Retired; Retired; Former warehouseman}   \\
    \cmidrule(r){1-4}
    Employment\_Other & The patient is not described with previous employment labels nor job. This entity describes other employment situations: unclear work status without temporality, disability, long sick leave, irregular situation, housewife, volunteer work, etc.). & \textit{Il a été courtier dans le textile ; Elle a travaillé comme dentiste ; Travailleur handicapé et ne travaille plus depuis 2011 ; En invalidité ; Mère au foyer}  & \textit{He was a textile broker; She worked as a dentist; Disabled worker and hasn't worked since 2011; On disability; Housewife}   \\
    \cmidrule(r){1-4}
    Alcohol & The alcohol use is annotated as an event. Only the trigger anchor is annotated here. The status of alcohol use is annotated through a relationship with the StatusTime entity. & \textit{Alcool; Ethylisme; CAD; OH; Intoxication ethylique}  & \textit{Alcohol; Ethylism; CAD; OH; Ethyl intoxication}   \\
    \cmidrule(r){1-4}
    Tobacco & The tobacco use is annotated as an event. Only the trigger anchor is annotated here. The status of tobacco use is annotated through a relationship with the StatusTime entity. & \textit{Tabac; Tabagisme ; Consommation tabagique}  & \textit{Tobacco; Smoking; Tobacco consumption}   \\
    \cmidrule(r){1-4}
    Drug & The drug use is annotated as an event. Only the trigger anchor is annotated here. The status of drug use is annotated through a relationship with the StatusTime entity. & \textit{Héroïne en fumette; Cannabis; drogue}  & \textit{Heroin smoking; Cannabis; Drugs}   \\
    \cmidrule(r){1-4}
    Housing\_Yes & The patient is described as having a housing stability. & \textit{Habite à Nantes; Pas d’aides à domicile ; Vit dans une maison}  & \textit{Lives in Nantes; No home help; Lives in a house}   \\
    \cmidrule(r){1-4}
    Housing\_No & The patient is described as having a housing instability. & \textit{Sans domicile fixe; Hébergé chez des tiers}  & \textit{Homeless; Living with a third party}   \\
    \cmidrule(r){1-4}
    PhysicalActivity\_Yes & The patient is described as exercising regularly. & \textit{Fait du vélo; Marche 30min par jour}  & \textit{Cycling; Walking 30min a day}   \\
    \cmidrule(r){1-4}
    PhysicalActivity\_No & The patient is described as not exercising regularly. & \textit{Marche avec une canne, sort peu ; Pas d’activité sportive; Sédentaire ; Performance status 3}  & \textit{Walks with a cane, goes out little; No sports activity; Sedentary; Performance status 3}   \\
    \cmidrule(r){1-4}
    Income & Any documentation about patient’s financial resources. It includes social aids. & \textit{RSA; AAH}  & \textit{Active solidarity income; Disabled adult allowance}   \\
    \cmidrule(r){1-4}
    Education & Any documentation about patient’s education. & \textit{Certificat d’étude; BTS Informatique; Licence en droit}  & \textit{Certificate of Study; Associate degree in Computer Science; Bachelor's Degree in Law}   \\
    \cmidrule(r){1-4}
    Ethnicity/Country of birth & The patient’s country of birth. & \textit{Originaire du Maroc; Originaire d’Érythrée}  & \textit{From Morocco ; From Eritrea}   \\
    StatusTime & The status of substance consumption (alcohol, tobacco and drug). An attribute is associated to this entity with possibles values: current, past, none. & \textit{Ne consomme pas d’alcool}  & \textit{No alcohol}   \\
    \cmidrule(r){1-4}
    History & Any date of occurrence of an event. & \textit{Il y a 8 ans; Il a arrêté depuis 3 ans}  & \textit{8 years ago; He stopped 3 years ago}   \\
    \cmidrule(r){1-4}
    Duration & Any exposure duration (to substance use, house instability, etc.). & \textit{Il a fumé pendant 20 ans.}  & \textit{He smoked for 20 years.}   \\
    \cmidrule(r){1-4}
    Amount & Any amount related to SDoH, such as Descendants or substance use. & \textit{20 paquets de cigarettes; 3 verres de vin; 2 enfants}  & \textit{20 packs of cigarettes; 3 glasses of wine; 2 children;}   \\
    \cmidrule(r){1-4}
    Frequency & Any frequency related to SDoH, such as physical activity or substance use. & \textit{Par jour; par mois; occasionnellement}  & \textit{Daily; monthly; occasionally}   \\
    \cmidrule(r){1-4}
    Type & Specify certain broad entities to get fine-grained details, such as the type of descendants, the type of drugs or alcohol used. & \textit{Enfants; petits-enfants; vin; bière ; cigarette ; cocaine ; cannabis}  & \textit{Children; grandchildren; wine; beer; cigarettes; cocaine; cannabis}   \\
\end{tabularx}
}
\setcounter{table}{1}

The annotation scheme also includes six relations (Table~\ref{tab:relations}):
\begin{itemize}
    \item Status: encodes the status (current, past, or none) of the substance use (alcohol, tobacco, or drugs).
    \item History: links any event to its date of occurrence
    \item Duration: encodes the duration of exposure to substance use
    \item Amount: encodes the quantity of substance use or the number of children in a lineage. Units of measurement: number of glasses, number of cigarettes, number of children, grams, etc.
    \item Frequency: gives the frequency of an event’s occurrence. This relation is also used when the amount related to substance use is not precise enough. For example: drinks occasionally
    \item Type: details certain entities, such as the type of lineage or the type of substance use.
\end{itemize}

\begin{table}[!ht]
  \caption{Relations. Involved entities in bold indicates the relations is required when the SDoH entity is annotated. * indicates that any entity can be linked.}
  \label{tab:relations}
  \vspace{6pt}
  \centering
{\small
    \begin{tabular}{p{0.25\columnwidth} p{0.30\columnwidth}}
    \toprule
    \textbf{Relation type}     & \textbf{Involved entities}  \\
    \midrule
    Status & \textbf{Tobacco – StatusTime}  \\
     & \textbf{Alcohol – StatusTime}  \\
     & \textbf{Drug – StatusTime}  \\
    \cmidrule(r){1-2}
    Amount & Descendants\_Yes – Amount  \\
     & Tobacco – Amount  \\
     & Alcohol – Amount  \\
     & Drug – Amount  \\
    \cmidrule(r){1-2}
    Duration & Tobacco – Amount  \\
     & Alcohol – Amount  \\
     & Drug – Amount  \\
    \cmidrule(r){1-2}
    Frequency & Physical\_Activity – Frequency  \\
     & Tobacco – Frequency  \\
     & Alcohol – Frequency  \\
     & Drug – Frequency  \\
    \cmidrule(r){1-2}
    History & * – History \\
    \cmidrule(r){1-2}
    Type & Descendants\_Yes – Type  \\
     & Tobacco – Type  \\
     & Alcohol – Type  \\
     & Drug – Type  \\
    \bottomrule
  \end{tabular}
  }
\end{table}

Following the work on SHAC corpus~\cite{lybarger2021annotating}, we annotated the substance use (tobacco, alcohol, and drugs) using an event-based scheme characterized by a trigger entity and status-related attributes.
We used the BRAT Rapid Annotation Tool (BRAT) for datasets annotation~\cite{stenetorp2012brat}. For the MUSCADET-InHouse dataset, the annotation process was carried out in three phases, with inter-annotator agreement calculated at the end of the first two phases: (1) a preliminary annotation phase on 100 documents by three annotators (PCDB, a physician; AB, an NLP researcher; and MK, an epidemiologist) to evaluate the annotation scheme and refine the guidelines; (2) a second annotation phase on 200 documents by two annotators (PCDB, AB) to validate the modifications made to the guidelines following the first phase; and (3) a final annotation phase during which each annotator worked independently according to the finalized annotation guidelines. 

For the MUSCADET-Synthetic dataset, all texts were annotated by three annotators (PCDB, AB, MK). The UW-FrenchSDOH dataset was annotated by a single annotator (AB), while the InHouse Tuberculosis and ALS dataset was annotated by two annotators (PCDB, AB). We computed inter-annotator agreement (IAA) values for entities using F-measure from the open-source tool bratiaa\footnote{\href{https://github.com/kldtz/bratiaa}{https://github.com/kldtz/bratiaa}} for each annotator pair. For relation annotations, F-measure scores were performed using an in-house script. The BRAT configuration files and the scripts for IAA computation are are available in the project's repository\footnote{\href{https://github.com/CliniqueDesDonnees/SDoH}{https://github.com/CliniqueDesDonnees/SDoH}}.

\subsection{Annotation Statistics}

Table~\ref{tab:entities_distrib} presents the distribution of entity types across all datasets. The most frequent entities are Living\_WithOthers, MaritalStatus\_InRelationship, Descendants\_Yes, Job, Tobacco, Alcohol and Housing\_Yes, while the remaining entities appear less frequently. Similarly, Table~\ref{tab:relations_distrib} presents the distribution of relation types across all datasets. The most common relations are Status, Amount, and Type, the latter two being highly associated with the entity Descendants\_Yes.

\begin{table}[!h]
  \caption{Entities distribution in all datasets. $n$ corresponds to the number of documents.}
  \label{tab:entities_distrib}
  \vspace{5pt}
  \centering
  \resizebox{\columnwidth}{!}{
\begin{tabular}{l|ccc|c|c|c}
    \toprule
    \textbf{Entities} 
    & \multicolumn{3}{c|}{\textbf{MUSCADET-InHouse}} 
    & \textbf{MUSCADET-Synthetic}
    & \textbf{UW-FrenchSDOH}
    & \textbf{InHouse Tuberculosis} \\

    &  &  &  & & & \textbf{and ALS} \\

    & \multicolumn{3}{c|}{($n=1{,}700$)}
    & ($n=340$)
    & ($n=364$)
    & ($n=400$) \\
    
    \cmidrule(lr){2-7}
     & \makebox[0.9cm]{\textit{Train}} & \makebox[0.9cm]{\textit{Dev}} & \makebox[0.9cm]{\textit{Test}}
     & \textit{Test} & \textit{Test} & \textit{Test} \\
    
    \midrule
    Living\_Alone & 194 & 32 & 55 & 37 & 17 & 10 \\
    Living\_WithOthers & 412 & 61 & 128 & 89 & 73 & 44 \\
    MaritalStatus\_Single & 50 & 9 & 14 & 20 & 20 & 6 \\
    MaritalStatus\_InRelationship & 629 & 72 & 184 & 132 & 127 & 72 \\
    MaritalStatus\_Divorced & 70 & 13 & 14 & 17 & 18 & 8 \\
    MaritalStatus\_Widowed & 69 & 8 & 17 & 18 & 9 & 3 \\
    Descendants\_Yes & 845 & 101 & 226 & 154 & 80 & 64 \\
    Descendants\_No & 98 & 13 & 25 & 33 & 5 & 6 \\
    Job	& 828 & 109 & 247 & 216 & 95 & 58 \\
    Last\_job & 751 & 100 & 217 & 204 & 85 & 49 \\
    Employment\_Working & 348 & 53 & 92 & 114 & 56 & 15 \\
    Employment\_Unemployed & 113 & 18 & 31 & 17 & 9 & 7 \\
    Employment\_Student & 30 & 4 & 14 & 17 & 8 & 0 \\
    Employment\_Pensioner & 305 & 45 & 91 & 48 & 34 & 23 \\
    Employment\_Other & 82 & 3 & 32 & 21 & 16 & 8 \\
    Alcohol & 498 & 70 & 127 & 193 & 251 & 42 \\
    Tobacco & 627 & 94 & 164 & 219 & 263 & 60 \\
    Drug & 78 & 11 & 20 & 101 & 136 & 17 \\
    Housing\_Yes & 682 & 86 & 203 & 122 & 48 & 137 \\
    Housing\_No	& 16 & 3 & 3 & 13 & 0 & 2 \\
    PhysicalActivity\_Yes & 154 & 23 & 33 & 63 & 21 & 4 \\
    PhysicalActivity\_No & 37 & 6 & 11 & 14 & 3 & 0 \\
    Income & 24 & 3 & 13 & 8 & 0 & 2 \\
    Education & 58 & 5 & 18 & 20 & 3 & 1 \\
    Ethnicity & 69 & 13 & 21 & 18 & 1 & 27 \\
    \bottomrule
    \end{tabular}
    }
\end{table}

\begin{table}[!h]
  \caption{Relations distribution in all datasets. $n$ corresponds to the number of documents.}
  \label{tab:relations_distrib}
  \vspace{5pt}
  \centering
  \resizebox{\columnwidth}{!}{
\begin{tabular}{l|ccc|c|c|c}
    \toprule
    \textbf{Relations} 
    & \multicolumn{3}{c|}{\textbf{MUSCADET-InHouse}} 
    & \textbf{MUSCADET-Synthetic}
    & \textbf{UW-FrenchSDOH}
    & \textbf{InHouse Tuberculosis} \\

    &  &  &  & & & \textbf{and ALS} \\

    & \multicolumn{3}{c|}{($n=1{,}700$)}
    & ($n=340$)
    & ($n=364$)
    & ($n=400$) \\
    
    \cmidrule(lr){2-7}
     & \makebox[0.9cm]{\textit{Train}} & \makebox[0.9cm]{\textit{Dev}} & \makebox[0.9cm]{\textit{Test}}
     & \textit{Test} & \textit{Test} & \textit{Test} \\
    
    \midrule
    Status & 1,206 & 175 & 314 & 514 & 646 & 117 \\
    Amount & 1,182 & 155 & 301 & 253 & 148 & 82 \\
    Duration & 51 & 8 & 18 & 12 & 33 & 10 \\
    Frequency & 489 & 68 & 124 & 124 & 135 & 35 \\
    History & 262 & 30 & 73 & 44 & 55 & 13 \\
    Type & 1,096 & 146 & 289 & 261 & 154 & 96 \\
    \bottomrule
    \end{tabular}
    }
\end{table}

\subsection{Experiment}

\begin{figure}[!h]
  \centering
  \includegraphics[width=0.9\textwidth]{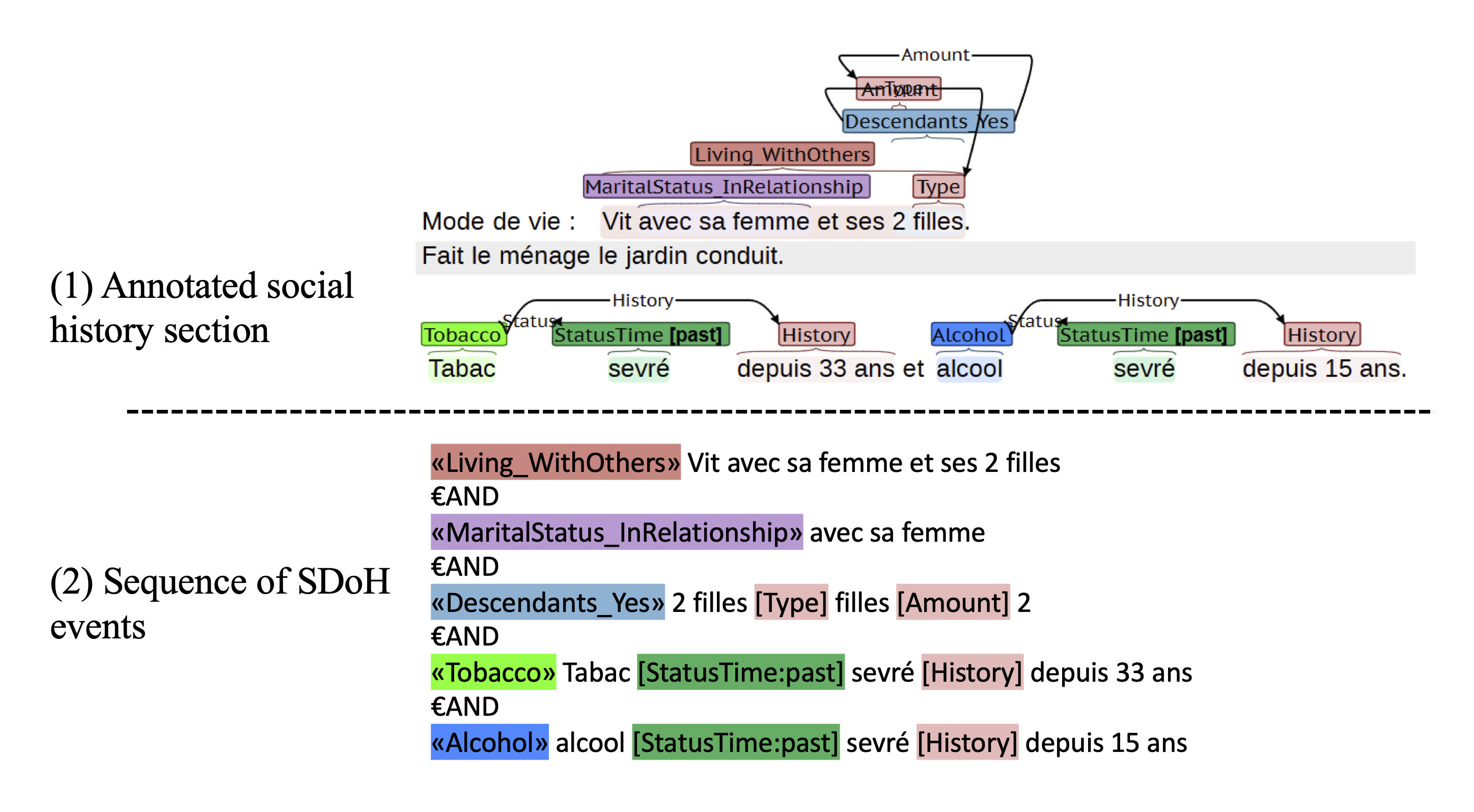}
  \caption{Example of (1) annotated social history section and (2) the corresponding structured sequence of SDoH events. Example translated into English: \textit{Social History: Lives with his wife and two daughters. Does the housework, gardening, and drives. Has been tobacco-free for 33 years and alcohol-free for 15 years}.}
  \label{fig:figure3}
\end{figure}

Following recent studies on SDoH extraction leveraging large language models (LLMs)~\cite{guevara2024large,romanowski2023extracting}, we used Flan-T5-Large model~\cite{chung2024scaling} in our experiments. We formulated SDoH extraction as a text-to-structure translation task, where the model receives a social history section from a clinical note and generates a linearized sequence of SDoH events. This sequence-to-sequence (seq2seq) formulation allows the model to jointly predict entities, attributes, and their relations within a single decoding pass. As illustrated in Figure~\ref{fig:figure3}, the output events are ordered from left to right according to their token offsets in the original text.
To train the model, Flan-T5-Large was fine-tuned on the MUSCADET-InHouse training set, with input-output representations presented in Figure~\ref{fig:figure3}. The model was fine-tuned for 10 epochs on two 24GB NVIDIA RTX 4090 GPUs. 

During inference for evaluation, the sequence of SDoH events generated by the model was post-processed to recover token offsets corresponding to entities and relations. The source text was then searched for tokens matching the SDoH events in the output sequence. However, the output sequence tokens often matched multiple offsets in the source text. Ambiguities were resolved by applying distinct strategies for entities and relations. For entities, when multiple non-overlapping matches were found, we selected the leftmost occurrence in the source text that had not already been extracted. For relations, we selected the match nearest to the associated entity to ensure contextual accuracy.
For the InHouse Tuberculosis and ALS dataset, we evaluated the model using both the full clinical notes and the social history sections alone after preprocessing to assess its robustness on non-social history section texts.

\subsection{Evaluation}

We conducted the evaluation at two levels: (i) SDoH factors and associated values, and (ii) fine-grained SDoH extraction including all entities and relations.

In the level 1 evaluation, we assessed the exact match presence of labeled entities in the gold standard and the model’s predictions. For alcohol, tobacco, and drug use, we included the corresponding Status relations to convert span-only entities into labeled entities (e.g., for Tobacco: Tobacco\_StatusTime:current, Tobacco\_StatusTime:past, Tobacco\_StatusTime:none).

\begin{figure}[!ht]
  \centering
  \includegraphics[width=0.65\textwidth]{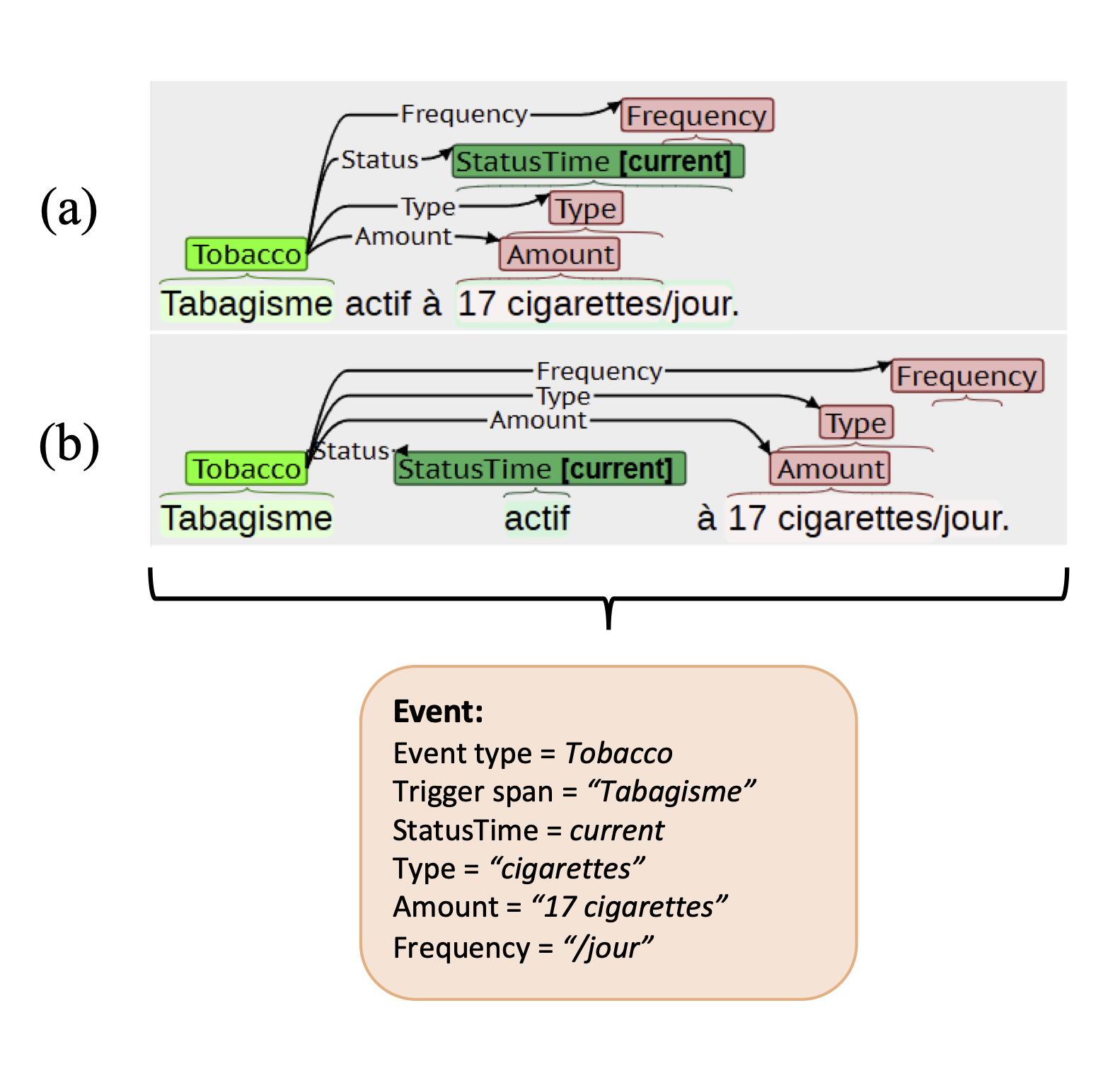}
  \caption{Examples of substance use annotated as events. Annotations (a) and (b) are considered equivalent. English translation of the example: \textit{Active smoking at 17 cigarettes per day.}}
  \label{fig:figure2}
\end{figure}

In the level 2 evaluation, we evaluated the extraction of SDoH as a slot-filling task, following prior work on evaluating SDoH extraction models in the context of 2022 n2c2/UW Shared Task~\cite{lybarger20232022}. This approach allows for multiple equivalent span annotations. Figure~\ref{fig:figure2} illustrates this by presenting the same sentence with two equivalent sets of annotations.
Event equivalence was defined using two criteria for model evaluation: exact-match spans and overlap-match spans. In the exact match setting, two events were considered equivalent if the entity offsets matched exactly between the gold standard and predictions, and their associated relation offsets also matched exactly. In the overlap match setting, two events were considered equivalent if the entity offsets shared at least one overlapping character between the gold standard and predictions, and their associated relation offsets also shared at least one overlapping character.
The performance of the model for all evaluation settings was measured using precision (P), recall (R), and F1-score (F1).

\subsection{Comparison with Structured EHR Data}

To assess the completeness of SDoH documentation in structured versus unstructured EHR data, we collected Z-codes for all patients in the MUSCADET-InHouse dataset. Z-codes are ICD-10 codes that describe factors that influence health status and healthcare utilization when the primary reason for the encounter is not a specific disease or injury, which partially include SDoH-related codes. All collected Z-codes for MUSCADET-InHouse patients were manually mapped to SDoH categories if relevant (see Supplementary Table~\ref{tab:zcodes_mapping}). We compared the presence of one or more SDoH categories in manually annotated text from MUSCADET-InHouse against the corresponding patient’s Z-codes from structured EHR data.

\section{Results}

\subsection{Inter-Annotator Agreement}

Table~\ref{tab:iaa} presents the inter-annotator agreement scores (F-measure). During the first phase of annotation for the MUSCADET-InHouse dataset, the average entity agreement was 0.689 before adjudication; it improved to 0.725 in the second phase. A similar trend was observed for relations, with an F-measure of 0.795 in the first phase, increasing to 0.829 in the second phase. For the MUSCADET-Synthetic dataset, the average agreement was 0.742 for entities and 0.788 for relations.

\begin{table}[!h]
  \caption{Inter-annotator agreement.}
  \label{tab:iaa}
  \vspace{5pt}
  \centering
{\small \begin{tabular}{l|c|c}
    \toprule
     
    & \textbf{Entities – F-Measure}
    & \textbf{Relations – F-Measure} \\
    \midrule

    \textbf{MUSCADET-InHouse}
    & 
    & \\

    Phase 1 – 100 documents
    & 0.689
    & 0.795 \\

    Phase 2 – 200 documents
    & 0.725
    & 0.829 \\

   \midrule 

    \textbf{MUSCADET-Synthetic}
    & 0.742
    & 0.788 \\
    \bottomrule
    \end{tabular}}
\end{table}

\subsection{Model Performance}

Table~\ref{tab:macro_score} shows the macro-averaged performance of the fine-tuned Flan-T5-Large model on all datasets. The model demonstrates stable performance when evaluated on the social history sections alone, achieving a macro-F1 score ranging from 0.7618 to 0.7863 in the level 1 evaluation setting (SDoH entities with associated values), and from 0.3934 to 0.4804 in the level 2 (SDoH extraction with all entities and relations) under the exact match criteria. However, when applied to full clinical notes from the InHouse Tuberculosis and ALS dataset, the model produces a high number of false positives, resulting in a substantial drop in performance, yielding a macro-F1 of 0.4017 in level 1, and 0.0451 in level 2 evaluation.
The model’s performance on the UW-FrenchSDOH dataset is slightly lower in the level 2 evaluation, likely due to the dataset being a translated dataset from English to French. Although the SDoH-related terms are accurately translated, the word order and writing style retain Anglophone patterns. Since the model was not trained on such translated or non-native-like data, it struggles to extract SDoH text spans precisely.

\begin{table}[!ht]
  \caption{Macro-averaged precision, recall, and F1 metrics of the seq2seq FlanT5-large model.}
  \vspace{5pt}
  \label{tab:macro_score}
  \centering
{\small \begin{tabular}{l|c|c|c}
    \toprule

    \textbf{Dataset} & \textbf{Precision} & \textbf{Recall} & \textbf{F1-score} \\
    \midrule

    \textbf{MUSCADET-InHouse} & & & \\
    SDoH entities with associated values (level 1) & 0.8133 & 0.7708 & 0.7863 \\
    All Entities and relations – Exact match (level 2) & 0.5145 & 0.4790 & 0.4804 \\
    All Entities and relations – Overlap match (level 2) & 0.6738 & 0.6313 & 0.6290 \\
   \midrule 

    \textbf{MUSCADET-Synthetic} & & & \\
    SDoH entities with associated values (level 1) & 0.8851 & 0.7344 & 0.7795 \\
    All Entities and relations – Exact match (level 2) & 0.5343 & 0.4429 & 0.4629 \\
    All Entities and relations – Overlap match (level 2) & 0.7222 & 0.5978 & 0.6297 \\
   \midrule 

    \textbf{UW-FrenchSDOH} & & & \\
    SDoH entities with associated values (level 1) & 0.7732 & 0.7972 & 0.7618 \\
    All Entities and relations – Exact match (level 2) & 0.3985 & 0.4788 & 0.3934 \\
    All Entities and relations – Overlap match (level 2) & 0.5716 & 0.6314 & 0.5446 \\
   \midrule 

    \textbf{InHouse Tuberculosis and ALS – Raw full texts} & & & \\
    SDoH entities with associated values (level 1) & 0.4167 & 0.5765 & 0.4017 \\
    All Entities and relations – Exact match (level 2) & 0.0471 & 0.0521 & 0.0451 \\
    All Entities and relations – Overlap match (level 2) & 0.1059 & 0.1518 & 0.1058 \\
   \midrule 

    \textbf{InHouse Tuberculosis and ALS – Social history sections only} & & & \\
    SDoH entities with associated values (level 1) & 0.8499 & 0.7487 & 0.7893 \\
    All Entities and relations – Exact match (level 2) & 0.4847 & 0.4325 & 0.4426 \\
    All Entities and relations – Overlap match (level 2) & 0.7316 & 0.6497 & 0.6700 \\
    
    \bottomrule
    \end{tabular}}
\end{table}

Table~\ref{tab:perf_sdoh} presents the model’s performance for each SDoH category. Some categories, such as living condition, marital status, descendants, job, tobacco and alcohol use, are well modeled by Flan-T5-Large with F1 scores over 0.80. These SDoH categories are often expressed in consistent, structured ways in clinical documents, making it easier for the model to learn their patterns. In contrast, categories such as employment status, housing, physical activity, income, and education present greater challenges. The model struggles to achieve consistent performance on these categories, likely due to the greater linguistic variability and contextual diversity in how these factors are documented in the clinical notes. This variability, along with the scarcity of certain SDoH categories, make generalization more difficult, especially when annotations vary in phrasing or context.

\subsection{Impact of Applying Model on Non-Social History Section Text}

Applying the model to entire clinical documents significantly increases the number of incorrect predictions (false positives). Specifically, the model's performance on full-text documents is considerably lower than when applied to social history sections only, with a macro-F1 of 0.4017 compared to 0.7893 in the level 1 evaluation setting. This discrepancy is expected, as the model was trained exclusively on social history sections and does not generalize effectively to other parts of the clinical notes.
While restricting inference to social history sections improves precision and overall performance, this approach risks missing important SDoH information that may appear elsewhere in the document. Thus, there is a trade-off between achieving high precision and ensuring comprehensive recall of patient-related social information.

To assess the extent of information missed under this constraint, we compared the number of SDoH annotations in the InHouse Tuberculosis and ALS dataset across full documents versus social history sections only. Across the full dataset, 665 annotations were identified, of which 461 (69.3\%) were located within the social history sections. This indicates that 204 annotations (30.7\%) lie outside these sections. Among these, 87 annotations occurred in documents that do not include a social history section, while the remaining 117 were found outside the social history section in documents that did include one. Notably, 81 of the 117 were redundant —i.e., SDoH categories that were already mentioned within the corresponding social history section. The remaining 36 annotations represented unique SDoH information not captured in the social history sections. These were primarily related to substance use (tobacco, alcohol, and drug use), commonly discussed in sections such as medical history or risk factors.
In total, restricting model inference to social history sections results in 123 missed unique SDoH annotations (87 from documents without a social history section, and 36 unique mentions from documents with one), accounting for approximately 18.5\% of all SDoH annotations in the InHouse Tuberculosis and ALS dataset.

\subsection{Error Analysis}

Supplementary Table~\ref{tab:error_type} provides an overview of the primary differences observed between the model outputs and the reference annotations. Through qualitative inspection, we categorized these discrepancies into eight distinct error types: (1) human annotation errors, (2) false positives, (3) false negatives, (4) difficulties in adhering to the structured output format, and (5) cases where the predicted text span was correct but the associated label was incorrect. Additional discrepancies labeled as errors were, in fact, not entirely incorrect; these resulted from (6) post-processing rules—for instance, when multiple identical text spans were present—or from (7) model predictions that differed from the ground truth annotation in terms of text spans but were nonetheless valid in the context of the slot-filling task. A small number of errors also stemmed from (8) limitations of the tokenizer, which did not support several French characters, such as “ \textit{ï} ”. This led the model to generate incorrect forms, such as producing “\textit{cocane}” instead of “\textit{cocaïne}”, thereby introducing errors in post-processing.

\subsection{Comparison with Structured EHR Data}

Manual annotation of the MUSCADET-InHouse dataset identified at least one SDoH category in 98.5\% of patients (1621/1646). In contrast, structured EHR data, based on Z-codes, captured SDoH information in only 2.8\% of cases (46/1646). Among these, 17 SDoH mentions overlapped between the two sources. The remaining non-overlapping instances from the structured data were mainly associated with Z-codes such as Z29.0 and Z60.20, which correspond to living alone.

\clearpage
\thispagestyle{empty}

\begin{table}[!h]
  \caption{Performance of Flan-T5-Large on SDoH categories.}
  \label{tab:perf_sdoh}
  \vspace{6pt}
  \centering
    \resizebox{\columnwidth}{!}{
\begin{tabular}{l|ccc|ccc|ccc}
    \toprule

    \multirow{3}{*}{\textbf{SDoH category}} 
     & \multicolumn{3}{c|}{\textbf{SDoH entities with associated values}} & \multicolumn{6}{c}{\textbf{All entities and relations (level 2)}} \\

    \cline{5-10} 

     & \multicolumn{3}{c|}{\textbf{(level 1)}} & \multicolumn{3}{c|}{\raisebox{-0.3ex}{\textbf{Exact match}}} & \multicolumn{3}{c}{\raisebox{-0.3ex}{\textbf{Overlap match}}} \\

    \cline{2-10} 
    
      & \raisebox{-0.4ex}{\makebox[1.5cm]{\textbf{P}}} & \raisebox{-0.4ex}{\makebox[1.5cm]{\textbf{R}}} & \raisebox{-0.4ex}{\makebox[1.5cm]{\textbf{F1}}} & \raisebox{-0.4ex}{\textbf{P}} & \raisebox{-0.4ex}{\textbf{R}} & \raisebox{-0.4ex}{\textbf{F1}} & \raisebox{-0.4ex}{\textbf{P}} & \raisebox{-0.3ex}{\textbf{R}} & \raisebox{-0.4ex}{\textbf{F1}} \\
    \midrule

    \textbf{Living condition} & & & & & & & & & \\
    MUSCADET-InHouse & 0.9141 & 0.9088 & 0.9113 & 0.8228 & 0.8088 & 0.8157 & 0.9101 & 0.8947 & 0.9023 \\
    MUSCADET-Synthetic & 0.8939 & 0.9112 & 0.9024 & 0.8439 & 0.8606 & 0.8521 & 0.8884 & 0.9056 & 0.8968 \\
    UW-FrenchSDOH & 0.8513 & 0.8727 & 0.8603 & 0.8422 & 0.8316 & 0.8308 & 0.8422 & 0.8316 & 0.8308 \\
    InHouse Tuberc./ALS – Raw full texts & 0.4861 & 0.6387 & 0.5265 & 0.1266 & 0.1568 & 0.1363 & 0.2873 & 0.3296 & 0.3007 \\
    InHouse Tuberc./ALS – Social history sections & 0.9630 & 0.6950 & 0.8073 & 0.8704 & 0.6274 & 0.7292 & 0.9445 & 0.6815 & 0.7917 \\
   
   \midrule 

    \textbf{Marital status} & & & & & & & & & \\
    MUSCADET-InHouse & 0.8681 & 0.9358 & 0.9007 & 0.6131 & 0.8089 & 0.6759 & 0.6701 & 0.8677 & 0.7337 \\
    MUSCADET-Synthetic & 0.9550 & 0.9206 & 0.9370 & 0.7300 & 0.7140 & 0.7211 & 0.8317 & 0.8213 & 0.8255 \\
    UW-FrenchSDOH & 0.9196 & 0.9115 & 0.9154 & 0.6647 & 0.5910 & 0.6138 & 0.7507 & 0.6783 & 0.7005 \\
    InHouse Tuberc./ALS – Raw full texts & 0.8385 & 0.5821 & 0.6778 & 0.0147 & 0.0104 & 0.0122 & 0.3199 & 0.2222 & 0.2587 \\
    InHouse Tuberc./ALS – Social history sections & 0.9500 & 0.7212 & 0.8131 & 0.8809 & 0.6330 & 0.7313 & 0.9341 & 0.6840 & 0.7834 \\

    \midrule

    \textbf{Descendants} & & & & & & & & & \\
    MUSCADET-InHouse & 0.9838 & 0.9891 & 0.9864 & 0.5769 & 0.6176 & 0.5963 & 0.8831 & 0.9371 & 0.9091 \\
    MUSCADET-Synthetic & 0.9733 & 0.9696 & 0.9715 & 0.6711 & 0.6730 & 0.6721 & 0.9548 & 0.9580 & 0.9564 \\
    UW-FrenchSDOH & 0.8114 & 0.9688 & 0.8831 & 0.4098 & 0.5210 & 0.4556 & 0.7221 & 0.8881 & 0.7919 \\
    InHouse Tuberc./ALS – Raw full texts & 0.7375 & 0.6919 & 0.6857 & 0.0886 & 0.0728 & 0.0784 & 0.2527 & 0.2359 & 0.2390 \\
    InHouse Tuberc/ALS – Social history sections & 0.9878 & 0.9045 & 0.9424 & 0.6814 & 0.6374 & 0.6574 & 0.9214 & 0.8756 & 0.8963 \\

    \midrule

    \textbf{Housing} & & & & & & & & & \\
    MUSCADET-InHouse & 0.4272 & 0.4147 & 0.4209 & 0.2158 & 0.2094 & 0.2125 & 0.3655 & 0.3547 & 0.3600 \\
    MUSCADET-Synthetic & 0.6286 & 0.4993 & 0.4824 & 0.3809 & 0.1983 & 0.2106 & 0.6091 & 0.4770 & 0.4615 \\
    UW-FrenchSDOH & 0.4265 & 0.6744 & 0.5225 & 0.1159 & 0.1667 & 0.1368 & 0.3623 & 0.5208 & 0.4274 \\
    InHouse Tuberc./ALS – Raw full texts & 0.1537 & 0.3281 & 0.2093 & 0.0156 & 0.0219 & 0.0182 & 0.0518 & 0.0730 & 0.0606 \\
    InHouse Tuberc./ALS – Social history sections & 0.4768 & 0.4184 & 0.4457 & 0.2322 & 0.1831 & 0.2047 & 0.4286 & 0.3381 & 0.3780 \\
    
    \midrule

    \textbf{Employment} & & & & & & & & & \\
    MUSCADET-InHouse & 0.8707 & 0.6892 & 0.7482 & 0.5945 & 0.5029 & 0.5387 & 0.7949 & 0.6282 & 0.6855 \\
    MUSCADET-Synthetic & 0.8248 & 0.6995 & 0.7275 & 0.6004 & 0.4370 & 0.4674 & 0.8158 & 0.5807 & 0.6339 \\
    UW-FrenchSDOH & 0.7138 & 0.7369 & 0.7054 & 0.6180 & 0.6042 & 0.5816 & 0.6368 & 0.6229 & 0.6018 \\
    InHouse Tuberc./ALS – Raw full texts & 0.1731 & 0.5366 & 0.2470 & 0.0332 & 0.0756 & 0.0447 & 0.1325 & 0.4086 & 0.1940 \\
    InHouse Tuberc./ALS – Social history sections & 0.8352 & 0.7303 & 0.7506 & 0.3728 & 0.3798 & 0.3744 & 0.8400 & 0.7138 & 0.7454 \\

    \midrule

    \textbf{Alcohol} & & & & & & & & & \\
    MUSCADET-InHouse & 0.8561 & 0.8169 & 0.8357 & 0.4472 & 0.4348 & 0.4405 & 0.7068 & 0.6834 & 0.6943 \\
    MUSCADET-Synthetic & 0.9257 & 0.6504 & 0.7443 & 0.4192 & 0.3318 & 0.3656 & 0.7526 & 0.5839 & 0.6490 \\
    UW-FrenchSDOH & 0.9379 & 0.8182 & 0.8708 & 0.0555 & 0.0662 & 0.0577 & 0.4988 & 0.4683 & 0.4763 \\
    InHouse Tuberc./ALS – Raw full texts & 0.3423 & 0.4998 & 0.3287 & 0.0563 & 0.0533 & 0.0497 & 0.1325 & 0.1584 & 0.1225 \\
    InHouse Tuberc./ALS – Social history sections & 0.9364 & 0.8714 & 0.9000 & 0.1777 & 0.1655 & 0.1708 & 0.6574 & 0.6205 & 0.6277 \\

    \midrule

    \textbf{Tobacco} & & & & & & & & & \\
    MUSCADET-InHouse & 0.9825 & 0.8888 & 0.9332 & 0.7019 & 0.6374 & 0.6663 & 0.8367 & 0.7599 & 0.7945 \\
    MUSCADET-Synthetic & 0.8944 & 0.7566 & 0.8181 & 0.6076 & 0.5358 & 0.5481 & 0.7310 & 0.6537 & 0.6853 \\
    UW-FrenchSDOH & 0.8664 & 0.8597 & 0.8607 & 0.4437 & 0.4122 & 0.4241 & 0.6566 & 0.6214 & 0.6343 \\
    InHouse Tuberc./ALS – Raw full texts & 0.5460 & 0.6102 & 0.5500 & 0.0414 & 0.0590 & 0.0475 & 0.0749 & 0.0911 & 0.0799 \\
    InHouse Tuberc./ALS – Social history sections & 1.0000 & 0.8975 & 0.9445 & 0.6893 & 0.6072 & 0.6379 & 0.9025 & 0.7890 & 0.8318 \\

    \midrule

    \textbf{Drug} & & & & & & & & & \\
    MUSCADET-InHouse & 0.6000 & 0.5048 & 0.5444 & 0.6945 & 0.4222 & 0.5126 & 0.7445 & 0.4556 & 0.5526 \\
    MUSCADET-Synthetic & 0.9097 & 0.5598 & 0.6663 & 0.3304 & 0.1439 & 0.1906 & 0.6220 & 0.2911 & 0.3728 \\
    UW-FrenchSDOH & 0.7211 & 0.6354 & 0.5716 & 0.2852 & 0.2245 & 0.2324 & 0.5378 & 0.3661 & 0.3938 \\
    InHouse Tuberc./ALS – Raw full texts & 0.4468 & 0.5516 & 0.3404 & 0.0000 & 0.0000 & 0.0000 & 0.0000 & 0.0000 & 0.0000 \\
    InHouse Tuberc./ALS – Social history sections & 1.0000 & 1.0000 & 1.0000 & 0.4167 & 0.2315 & 0.2778 & 0.6667 & 0.4537 & 0.5093 \\

    \midrule
    
    \textbf{Physical activity} & & & & & & & & & \\
    MUSCADET-InHouse & 0.8167 & 0.7891 & 0.7955 & 0.1179 & 0.2689 & 0.1514 & 0.2504 & 0.5531 & 0.3177 \\
    MUSCADET-Synthetic & 0.9606 & 0.6429 & 0.7660 & 0.2249 & 0.1044 & 0.1374 & 0.3713 & 0.1996 & 0.2528 \\
    UW-FrenchSDOH & 0.7106 & 0.6970 & 0.6667 & 0.2302 & 0.1052 & 0.1302 & 0.6111 & 0.2083 & 0.2730 \\
    InHouse Tuberc./ALS – Raw full texts & 0.0259 & 0.7500 & 0.0500 & 0.0000 & 0.0000 & 0.0000 & 0.0046 & 0.1667 & 0.0089 \\
    InHouse Tuberc./ALS – Social history sections & 0.5000 & 0.5000 & 0.5000 & 0.0000 & 0.0000 & 0.0000 & 0.8333 & 0.8333 & 0.8333 \\

    \midrule

    \textbf{Job} & & & & & & & & & \\
    MUSCADET-InHouse & - & - & - & 0.7456 & 0.6883 & 0.7158	 & .9342 & 0.8623 & 0.8968 \\
    MUSCADET-Synthetic & - & - & - & 0.8010 & 0.7454 & 0.7722 & 0.8856 & 0.8241 & 0.8537 \\
    UW-FrenchSDOH & - & - & - & 0.6702 & 0.6632 & 0.6667 & 0.8191 & 0.8105 & 0.8148 \\
    InHouse Tuberc./ALS – Raw full texts & - & - & - & 0.0379 & 0.0862 & 0.0526 & 0.1061 & 0.2414 & 0.1474 \\
    InHouse Tuberc./ALS – Social history sections & - & - & - & 0.6667 & 0.5833 & 0.6222 & 0.8810 & 0.7708 & 0.8222 \\

    \midrule

    \textbf{Last job} & & & & & & & & & \\
    MUSCADET-InHouse & - & - & - & 0.701 & 0.659 & 0.6793 & 0.8676 & 0.8157 & 0.8409 \\
    MUSCADET-Synthetic & - & - & - & 0.7919 & 0.7647 & 0.7781 & 0.8680 & 0.8382 & 0.8529 \\
    UW-FrenchSDOH & - & - & - & 0.6818 & 0.7059 & 0.6936 & 0.8295 & 0.8588 & 0.8439 \\
    InHouse Tuberc./ALS – Raw full texts & - & - & - & 0.0315 & 0.0816 & 0.0455 & 0.0866 & 0.2245 & 0.1250 \\
    InHouse Tuberc./ALS – Social history sections & - & - & - & 0.6316 & 0.5854 & 0.6076 & 0.8158 & 0.7561 & 0.7848 \\

    \midrule

    \textbf{Income} & & & & & & & & & \\
    MUSCADET-InHouse & - & - & - & 0.2500 & 0.0769 & 0.1177 & 0.3750 & 0.1154 & 0.1765 \\
    MUSCADET-Synthetic & - & - & - & 0.5000 & 0.1250 & 0.2000 & 0.5000 & 0.1250 & 0.2000 \\
    UW-FrenchSDOH & - & - & - & - & - & - & - & - & - \\
    InHouse Tuberc./ALS – Raw full texts & - & - & - & 0.0000 & 0.0000 & 0.0000 & 0.0000 & 0.0000 & 0.0000 \\
    InHouse Tuberc./ALS – Social history sections & - & - & - & 0.0000 & 0.0000 & 0.0000 & 0.0000 & 0.0000 & 0.0000 \\

    \midrule
    
    \textbf{Education} & & & & & & & & & \\
    MUSCADET-InHouse & - & - & - & 0.3889 & 0.3889 & 0.3889 & 0.6111 & 0.6111 & 0.6111 \\
    MUSCADET-Synthetic & - & - & - & 0.3478 & 0.4000 & 0.3721 & 0.4348 & 0.5000 & 0.4651 \\
    UW-FrenchSDOH & - & - & - & 0.0526 & 0.3333 & 0.0909 & 0.0526 & 0.3333 & 0.0909 \\
    InHouse Tuberc./ALS – Raw full texts & - & - & - & 0.0000 & 0.0000 & 0.0000 & 0.0000 & 0.0000 & 0.0000 \\
    InHouse Tuberc./ALS – Social history sections & - & - & - & 0.5000 & 1.0000 & 0.6667 & 0.5000 & 1.0000 & 0.6667 \\

    \midrule

    \textbf{Ethnicity/Country of birth} & & & & & & & & & \\
    MUSCADET-InHouse & - & - & - & 0.7500 & 0.5714 & 0.6486 & 0.9375 & 0.7143 & 0.8108 \\
    MUSCADET-Synthetic & - & - & - & 0.2308 & 0.1667 & 0.1935 & 0.8462 & 0.6111 & 0.7097 \\
    UW-FrenchSDOH & - & - & - & 0.1111 & 1.0000 & 0.2000 & 0.1111 & 1.0000 & 0.2000 \\
    InHouse Tuberc./ALS – Raw full texts & - & - & - & 0.2143 & 0.1111 & 0.1463 & 0.4286 & 0.2222 & 0.2927 \\
    InHouse Tuberc./ALS – Social history sections & - & - & - & 0.6667 & 0.4211 & 0.5161 & 0.9167 & 0.5789 & 0.7097 \\
        
    \bottomrule
    \end{tabular}
    }
\end{table}

\clearpage

\section{Discussion}

We developed a sequence-to-sequence model to extract 13 SDoH categories from French clinical notes, demonstrating the potential of large language models for enhancing the collection of real-world SDoH data. Our model performed well in identifying SDoH mentions in clinical notes and showed consistent performance across four datasets, including two that are publicly available to the research community. SDoH mentions extracted from clinical notes identified 95.8\% patients with relevant information, compared to 2.8\% for ICD-10 codes from structured EHR data, underscoring the added value of unstructured data.
These results highlight the effectiveness of NLP approaches in leveraging unstructured clinical notes to improve the completeness of real-world data, which is often missing or sparsely represented in structured EHR data. For example, ICD-10 Z-codes describing SDoH (e.g. 'Problems relating to housing and economic circumstances') are used in less than 5\% of cases by clinicians in routine discharge coding practice, whereas automated NLP systems can recover comparable information with far less effort, requiring about one day of processing versus nine person-days per physician~\cite{gauthier2022automating}. Yet clinician documentation habits remain a bottleneck: in a U.S. study of $>$5 million patients, structured data such as address and race were well-documented, while housing, income, and social isolation were mentioned in less than 5\% of records~\cite{hatef2019assessing}. Greater clinician awareness and consistent recording of these factors are therefore essential.

Our model achieved strong performance (macro-F1 > 0.80) in identifying well-documented SDoH categories (living condition, marital status, descendants, smoking status, alcohol use, employment and physical activity) but lower scores for housing status and drug use. These discrepancies were primarily due to inconsistencies in human annotation, limited training data, and highly variable language, ranging from direct mentions (e.g., “apartment,” “house”) to more indirect or context-dependent references (e.g., “nursing home,” “home nurse,” “in-home assistance”). These results highlight the strengths of our approach in extracting high-level SDoH categories, which is particularly relevant for secondary use applications. Since the output is already structured for each SDoH category, it can be directly integrated into clinical databases and research cohorts without requiring additional post-processing. However, when more granular detail is needed, fine-grained SDoH extraction with entities and relations involves additional post-processing steps and result in lower and less stable performance across SDoH categories.

Our error analysis revealed several limitations in using language models to extract SDoH from French clinical notes. While such models are generally capable of identifying the presence of relevant concepts, they often struggle to precisely extract detailed information. This performance gap may be explained by multiple factors, including the relatively small number of models’ parameters compared to state-of-the-art architectures, and quality issues in the annotated data. Indeed, model performance is inherently limited by the quality and consistency of the annotations, which are challenging in the SDoH domain due to its conceptual complexity. Annotator bias and inconsistency further reduce reliability and, consequently, model accuracy.
Additional errors stem from the use of English-based tokenizers, which often mishandle accented characters. As a result, post-processing becomes difficult: the predicted spans cannot be reliably aligned with gold annotations, and character offsets are often miscalculated during evaluation. These issues underscore the need for tokenizers and models tailored to specific languages, as most publicly available models are English-centric and may not generalize well to other languages or multilingual contexts~\cite{zhang-etal-2023-dont,wendler2024llamas}.
Moreover, the generation-based approach introduces alignment errors during post-processing. Specifically, selecting the leftmost matching text span to align the generated SDoH outputs can result in incorrect mappings. Similarly, associating predicted entities with their nearest potential relation arguments can introduce a proximity bias, potentially overlooking longer-range dependencies.

Applying NLP to EHR data poses a persistent challenge of transferability. Models often struggle to maintain consistent performance across different patient sub-populations within the same institution, and even more so across hospitals or over time as clinical language and practices evolve. This limits the reliability, equity, and generalizability of NLP-driven insights, underscoring the importance of adaptable, continuously validated models in clinical settings. This highlights the need for ongoing ad-hoc validation studies, underlying the importance of methodological transparency in studies like ours, including the release of code and data.

Another key challenge in SDoH research is the scarcity of resources in languages other than English, which limits the development of NLP methods that account for social and cultural variability across healthcare systems. SDoH are deeply context-dependent, shaped by language, culture, policy, and local healthcare practice, making it essential to develop corpora that reflect diverse populations~\cite{faisal-anastasopoulos-2023-geographic}. A major motivation behind our work is to address this gap by providing a French-language SDoH corpus that is openly accessible and free from legal constraints. Given the sensitivity of medical data under the General Data Protection Regulation (GDPR), we adopted a dual approach to ensure compliance: generating synthetic social history sections authored by a physician and translating publicly available English-language a dataset from the University of Washington into French. This approach enables us to uphold privacy standards while advancing FAIR research principles. By introducing a corpus tailored to the French clinical context, we aim to promote inclusivity and facilitate the development of NLP methods for French-speaking populations and foster multilingual research in SDoH extraction.

Our study has several limitations that affect the generalizability of our findings. First, our training dataset was derived from a predominantly Caucasian population treated at one comprehensive center, Nantes University hospital. This demographic skew impacted certain SDoH categories, such as ethnicity, which are more likely to be documented for non-caucasian individuals. In addition, ethnic data are not usually collected by French physicians unless deemed relevant for healthcare purposes. In general, we observed variation in the amount of SDoH information available across populations. On average, patients born in France (n = 1397, mean = 5.45 (SD 2.32) SDoH mentions) had more SDoH information recorded (P < 0.01 using a Student’s t-test) than patients born outside France (n = 239, mean = 4.87 (SD 2.29) SDoH mentions), suggesting possible disparities in data completeness. A second limitation is that we trained our model only on the social history sections of clinical notes, rather than on full-text documents. While this decision reduced annotation effort, it limited the model’s ability to generalize to other sections. As a result, the model is not directly applicable to raw clinical notes and requires a preprocessing step to isolate the relevant sections prior to inference. This design choice may also reduce recall, as SDoH can also appear in other sections of clinical notes. Furthermore, because not all clinical notes include a social history section, information may be missed for certain patients.

The availability and quality of SDoH documentation in EHRs is often limited and inconsistent. Real-world data are primarily collected for clinical and administrative purposes by physicians during patient care, rather than for secondary use in research. Consequently, certain SDoH categories, apart from substance use, which is well known as a risk factor, are often overlooked during consultations. Several factors contribute to this under-documentation: lack of awareness among healthcare providers about the relevance of social factors to health outcomes, discomfort with asking about these factors, and restricted resources, staffing, and time to conduct screenings, which often compete with medical priorities~\cite{park2024factors}. Moreover, as Nantes University Hospital serves as the comprehensive center in our region, physicians there tend to focus more on medical care and less on the social environment than general practitioners~\cite{loo2025implementing}. As a result, SDoH information is frequently missing or incomplete, even in unstructured formats within the EHR. This under-documentation limits the ability to study social determinants at scale and hinders efforts to reduce health disparities. It also impairs the capacity of health systems to implement targeted, equity-oriented interventions based on complete, representative patient data.

\section{Conclusion}

Social determinants of health have a profound impact on both individual and population health outcomes, influencing morbidity, mortality, and healthcare access. Yet, SDoH are often under-documented in EHRs, particularly in structured data. In this work, we developed and evaluated a sequence-to-sequence language model to extract 13 SDoH categories from French clinical notes, demonstrating the effectiveness of NLP for improving the completeness of real-world health data. Our model consistently outperformed structured EHR data by identifying the majority of relevant SDoH across all patients and showed robust performance across multiple datasets, including publicly available benchmarks.

Future work will explore data augmentation techniques and the use of synthetic clinical text to improve the model’s generalization, and facilitate the open-source release of both the model and annotated training dataset to support reproducibility and multilingual SDoH research. Ultimately, advancing automated SDoH extraction from unstructured clinical text can support more equitable healthcare by enabling richer, more representative data for research, policy-making, and population health interventions.

\newpage

\section*{Acknowledgments}

This work was financially supported, in part, by the Agence Nationale de la Recherche (ANR) AIBy4 under contract ANR-20-THIA-0011, ANR MALADES under contract ANR-23-IAS1-0005, a grant from the French Ministry of Health (DAtAE2023-15744432) and the cluster DELPHI - NExT under contract ANR-16-IDEX-0007, integrated to France 2030 plan, by Région Pays de la Loire and by Nantes Métropole. We would also like to thank Emilie Varey for her administrative support in managing the project.

\section*{Authors Contribution}

AB: Conceptualization, Methodology (model training and evaluation), Investigation, Resources, Data curation, Writing- Original draft, Formal analysis, Visualization. PCDB: Conceptualization, Investigation, Resources, Data curation, Writing- Original draft preparation, Formal analysis, Visualization, Funding acquisition. MH: Resources, Methodology, Investigation, Formal analysis. RB: Supervision, Writing- Reviewing and Editing. EM: Supervision, Writing- Reviewing and Editing. RD: Supervision, Writing- Reviewing and Editing. BD: Supervision, Writing- Reviewing and Editing. PAG: Supervision, Writing- Reviewing and Editing. MK: Conceptualization, Methodology, Investigation, Writing- Reviewing and Editing, Project administration, Supervision.

\section*{Conflict of Interest Disclosure}

P.A. Gourraud is the founder of Methodomics (2008, www.methodomics.com) and of its spin-off Big Data Santé (2018, commercial name: “Octopize”). He acts as a consultant and/or contributor for several pharmaceutical and medical device companies. All related activities are conducted under institutional (university or hospital) contracts with the following entities: AstraZeneca, Amgen, Biogen, Boston Scientific, Cemka, Cook, Docaposte/Heva, Edimark, Ellipses, Elsevier, Grunenthal, Janssen, IAGE, Lek, Methodomics, Merck, Mérieux, Novartis, Octopize, Sanofi-Genzyme, Lifen, TuneInsight, and Aspire UAE. He serves as an unpaid board member of AXA mutual insurance (since 2021). He does not prescribe medications or medical devices and receives no personal remuneration. All other authors declare no competing interests.

\bibliographystyle{unsrt}
\bibliography{bibtex}






\newpage

\appendix

\section{Supplemental Material}

\begin{table}[!h]
  \caption{Z-codes mapping to SDoH categories.}
  \label{tab:zcodes_mapping}
  \vspace{6pt}
  \centering
\begin{tabular}{l|p{9cm}}
    \toprule
     
    \textbf{SDoH categories}
    & \textbf{Z-codes} \\
    
    \midrule

    Alcohol (current, past) & Z502 (alcohol withdrawal), Z714 (advice and monitoring for alcoholism) \\

    \midrule

    Tobacco (current) & Z720 (smoking-related difficulties) \\

    \midrule

    Drug & Z503 (rehabilitation for drug addicts and after drug abuse) \\

    \midrule

    Living condition (alone) & Z290 (isolation), Z602 (difficulties linked to loneliness) \\

    \midrule

    Descendants (yes) & Z370 (single birth, living child), Z372 (twin birth, twins born alive), Z391 (breastfeeding care and examinations), Z392 (routine postpartum check-up) \\
    
    \midrule

    Housing (no) & Z590 (difficulties associated with homelessness), Z598 (other difficulties related to housing and economic conditions) \\

    \midrule

    Ethnicity/Country of birth & Z603 (acculturation difficulties) \\
    
    \bottomrule
    \end{tabular}
    
\end{table}

\begin{table}[!h]
  \caption{Most common types of discrepancies between ground truth and model predictions.}
  \label{tab:error_type}
  \vspace{6pt}
  \centering
\begin{tabular}{l|l}
    \toprule
     
    \textbf{Error type}
    & \textbf{Count} \\
    
    \midrule

    Human annotation errors & 105 \\

    \midrule

    False negatives & 224 \\

    \midrule

    False positives & 125\\

    \midrule

    Output not structured as specified during model training & 63 \\

    \midrule

    Incorrect labels & 58 \\
    
    \midrule

    Postprocessing error (multiple matches in the source text) & 26 \\

    \midrule

    Correct label but different text spans & 94 \\

    \midrule

    Tokenization-related errors & 3 \\
    
    \bottomrule
    \end{tabular}
\end{table}

\end{document}